\RequirePackage{fix-cm}
\documentclass[twocolumn]{svjour3}          
\smartqed  
\usepackage{ulem}
\usepackage{graphicx}
\usepackage{epstopdf}
\usepackage{amsmath,amssymb}
\usepackage{color}
\usepackage{multirow}
\usepackage{subfig}
\usepackage{subfiles}
\usepackage{etoolbox}
\usepackage{enumerate}
\usepackage[switch]{lineno}
\usepackage{mathptmx}      
\newcommand{\revised}[1]{{\color{darkblue}#1}}

\definecolor{darkblue}{RGB}{0,0,0}

\begin{document}

\title{Learning Color Space Adaptation from Synthetic to Real Images of Cirrus Clouds
}


\author{Qing Lyu         \and
        Minghao Chen     \and
        Xiang Chen 
}


\institute{Q. Lyu \at
               the State Key Lab of CAD\&CG, Zijingang Campus, Zhejiang University, Hangzhou, Zhejiang, China 310058 \\
              \email{lyuqing@zju.edu.cn}           
           \and
           M. Chen \at
               the State Key Lab of CAD\&CG, Zijingang Campus, Zhejiang University, Hangzhou, Zhejiang, China 310058 \\
               \email{minghaochen01@gmail.com}
           \and
           X. Chen \at
              the State Key Lab of CAD\&CG, Zijingang Campus, Zhejiang University, Hangzhou, Zhejiang, China 310058 \\
               \email{xchen.cs@gmail.com}
}

\date{Received: date / Accepted: date}

\maketitle

\begin{abstract}
\revised{Cloud segmentation plays a crucial role in image analysis for climate modeling. Manually labeling the training data for cloud segmentation is time-consuming and error-prone. We explore to train segmentation networks with synthetic data due to the natural acquisition of pixel-level labels. Nevertheless, the domain gap between synthetic and real images significantly degrades the performance of the trained model. We propose a color space adaptation method to bridge the gap, by training a color-sensitive generator and discriminator to adapt synthetic data to real images in color space. Instead of transforming images by general convolutional kernels, we adopt a set of closed-form operations to make color-space adjustments while preserving the labels. We also construct a synthetic-to-real cirrus cloud dataset SynCloud and demonstrate the adaptation efficacy on the semantic segmentation task of cirrus clouds. With our adapted synthetic data for training the semantic segmentation, we achieve an improvement of $6.59\%$ when applied to real images, superior to alternative methods.}

\end{abstract}

\vspace{1\baselineskip}

\section{Introduction}\label{intro}

\vspace{-0.7\baselineskip}

\begin{figure*}[!t]
\centering
\subfloat[rendering images]{\includegraphics[width=.33\linewidth]{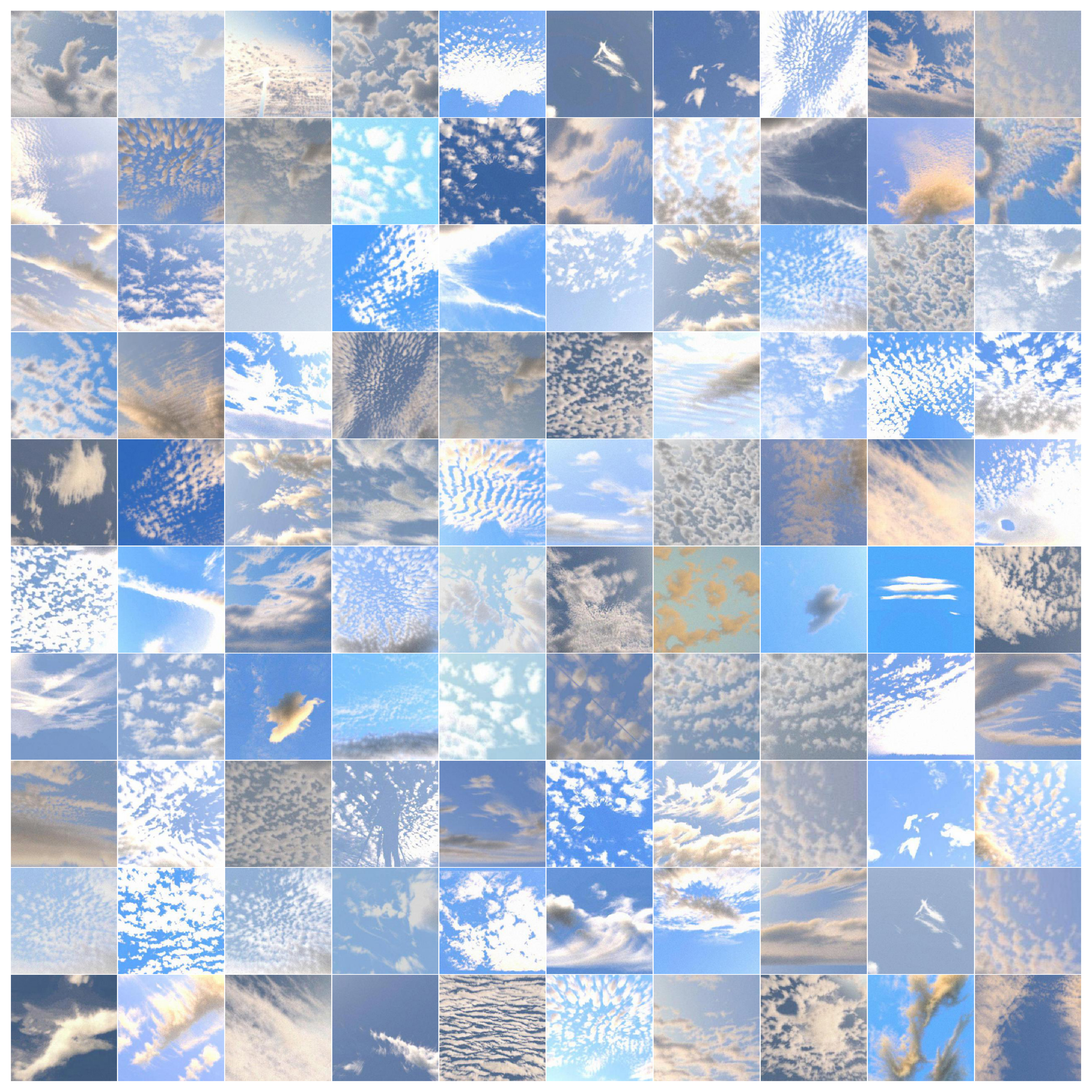}}
\subfloat[real images]{\includegraphics[width=.33\linewidth]{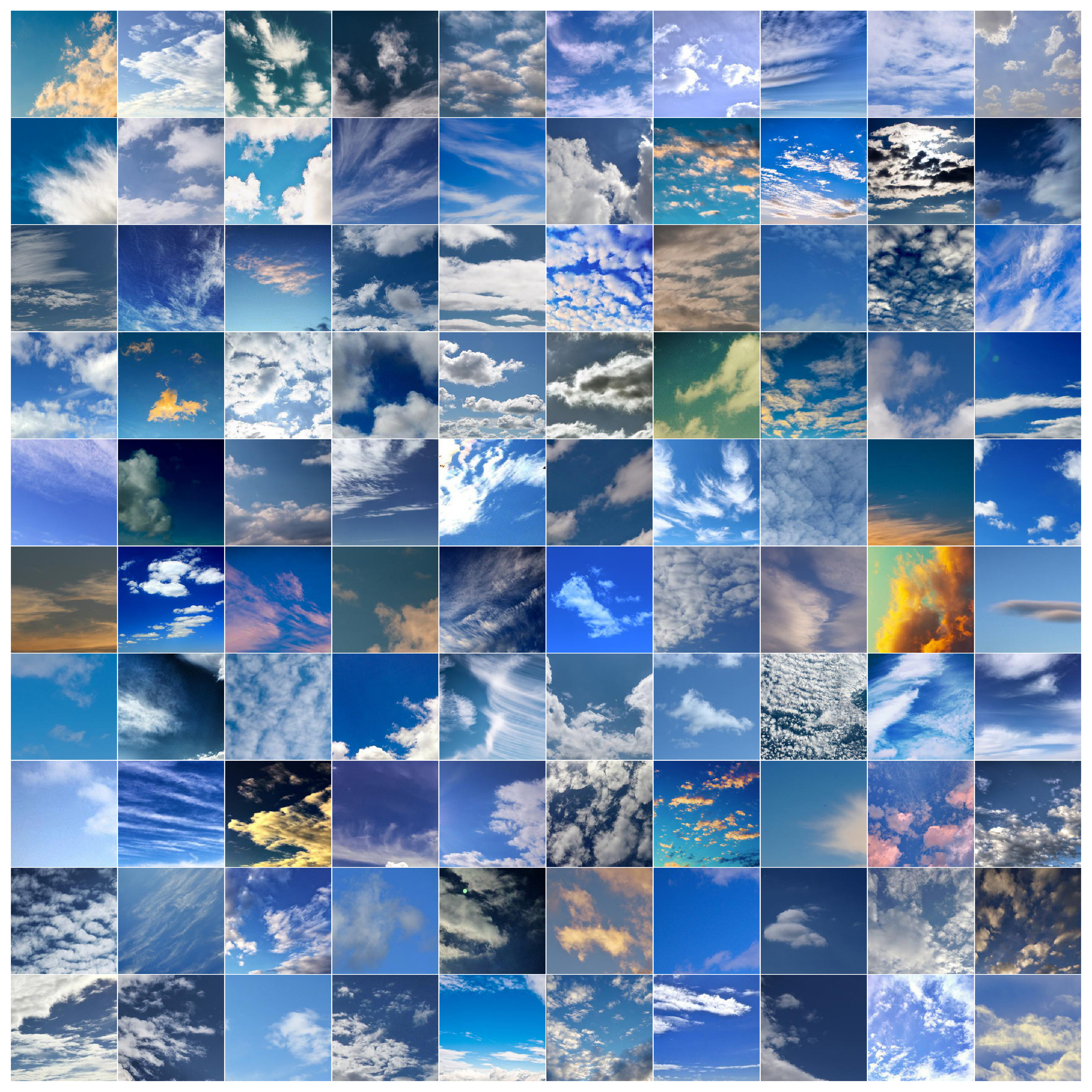}}
\subfloat[transferred images]{\includegraphics[width=.33\linewidth]{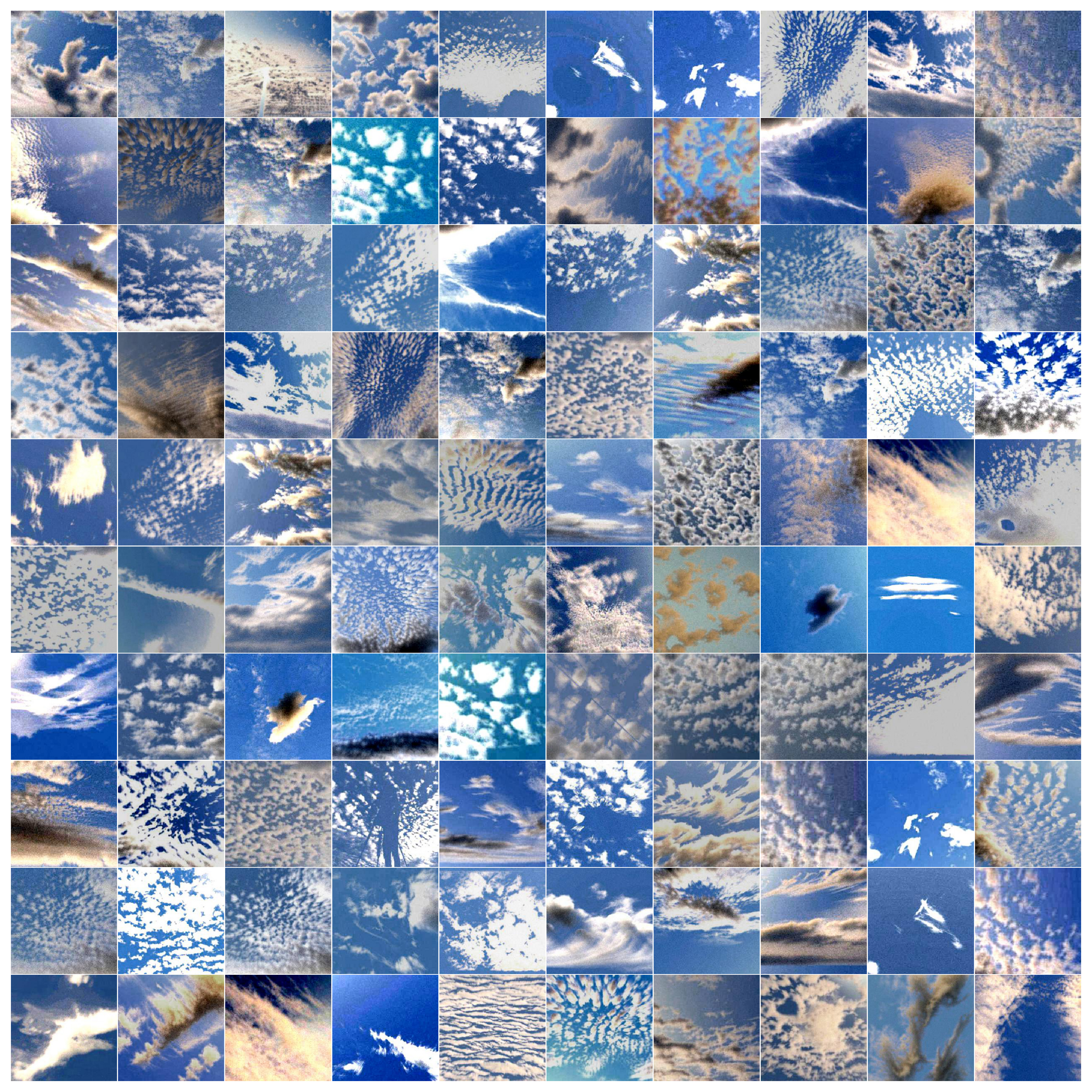}}\\
\caption{\textsc{Color Space Adaptation.} Compared with the original photo-realistic rendering images (a), the transferred images (c) after the color space adaption are visually more consistent with the real images (b). \revised{ Here we only present rendering images of similar colors. The full set of rendering images is diverse with a wide color range. Please refer to the Appendix for more results. }}
\label{fig:adap_results}
\end{figure*}

\revised{Cloud images are widely used in climate modeling, weather prediction, renewable energy generation, and satellite communications \cite{yuan2014comparison1,christodoulou2003multifeature,yuan2014comparison2,mahrooghy2012on}. Digital analysis of clouds and their features is necessary for these subjects. One of the first steps in cloud image analysis is cloud segmentation. Previous learning-based cloud segmentation methods are supervised and require a large number of training images with manually labeled ground-truth \cite{dev2017color,dianne2019deep,dev2019multi,dev2015multi}. Since manual labeling is time-consuming and error-prone, we explore to train cloud segmentation networks with synthetic images.}

Training on synthetic images has become increasingly popular in vision tasks, such as object detection\cite{sun2014from,massa2016deep,gaidon2016virtualworlds,tobin2017domain}, viewpoints estimation\cite{liebelt2010multi,stark2010back,su2015render,grabner20183d}, and semantic segmentation\cite{richter2016playing}. For example, photo-realistic rendering on 3D models can provide an accurate pixel-level annotation for each object in the scene, which eliminates the labeling cost for image segmentation learning. Under such circumstances, several synthetic datasets like Virtual KITTI \cite{gaidon2016virtualworlds} and SYNTHIA \cite{ros2016the} have been generated for training and evaluating vision models. When directly applying the synthetic dataset for real-world image tasks, performance degradation arises from the inherent domain gap. Such gaps may result from many reasons, {\it e.g.,} differences in geometric details, textures, backgrounds, and lightings. \revised{Some versatile domain transfer models are proposed to transfer synthetic images to real images \cite{bousmalis2017unsupervised,hoffman2018cycada}. However, the methods often break the preservation of original labels, especially for the pixel-level prediction tasks like semantic segmentation. Though there is a dedicated loss formulation for the preservation of images contents, segmentation labels cannot be preserved entirely due to the loss terms balancing.

We propose another strategy to solve the problem. Instead of loss design, we constrain the adaptation in color space. It would never destroy the segmentation label, which is crucial for tasks requiring accurate label-preserving. The reasons for choosing to adapt synthetic data in color space are two folds. First, we think that dividing the factors like color, shape, and textures provides a better opportunity for optimizing each subtask. Through division, we prevent the shape from changing and preserve original segmentation labels. Second, among other factors, the difference in color space plays a vital role. Since the synthetic data are produced from a photo-realistic 3D renderer, their textures are similar to real images. However, the synthetic data are raw images, while the real images are often post-processed either by digital cameras or photography software, which leads to a large discrepancy in their color space. Therefore we focus on the adaptation in color space.}

In this paper, we propose a color space adaptation framework to transfer a synthetic dataset to a given target of real images. Specifically, we adopt a set of basic operations for color space adjustments. These closed-form operations merely affect color representation without incurring any perceptual difference and label damage. We leverage such properties to generate sufficient adversarial examples for pretraining an effective classifier that is only sensitive to colors, but not to shapes and textures. \revised{Using the pre-trained color-sensitive classifier to evaluate the difference of color representations, we can train a generative network to adapt the synthetic dataset to the real images. Furthermore, we build a synthetic-to-real cirrus cloud dataset SynCloud to quantitatively evaluate the efficacy of the proposed color space adaptation framework for cloud image segmentation. The SynCloud dataset is consists of synthetic cloud images and unlabeled real images for adaptation training. The segmentation labels for synthetic images are automatically generated, while the labels of real images are manually annotated. All synthetic images are produced by physically-based rendering and are thus photorealistic. Fig.~\ref{fig:adap_results}-(a) displays some exemplars of our synthetic images. Our experiments show that the transferred dataset has significantly improved the segmentation performance compared with the original one. We also demonstrate some other applications of the cirrus cloud segmentation results, including 3D reconstruction, matting, and composition, for some interesting extensions.}

\revised{To summarize, the main contributions of our work are:
\begin{enumerate}
   \item We propose a simple yet robust framework for color space adaptation from synthetic data to real images while preserving the pixel-level labels.
   \item We construct a synthetic-to-real cirrus cloud dataset SynCloud to validate the efficacy of our adaptation method.
   \item We apply our method to tasks like style transfer and photo post-processing to show its potential.
\end{enumerate}
}

\vspace{-1\baselineskip}

\section{RELATED WORK}

\vspace{-0.7\baselineskip}

\revised{\paragraph{\textbf{Generative Adversarial Networks.}} Since proposed by Goodfellow {\it et al.} \cite{goodfellow2014generative} in 2014, generative adversarial networks (GANs) have attracted many research interests and achieved success in various image generation tasks such as the image super-resolution \cite{Ledig2017photo,Snderby2017amortised}. DCGANs \cite{radford2016unsupervised} embed convolutional networks (CNNs) in the GANs framework for image representation learning. Recently, researches have been exploring the possibility of applying GANs to the domain transfer problems like the image-to-image translation \cite{isola2017image,zhu2017unpaired,Taigman2017unsupervised,liu2016coupled,Park2019semantic}. Zhu {\it et al.} \cite{zhu2017unpaired} proposes CycleGAN with a novel cycle-consistent constraint for unpaired image-to-image translation. CoGAN \cite{liu2016coupled} trains two coupled GANs to synthesize pairs of corresponding images. With the development of GANs, domain adaptation from text to the image has also been explored \cite{Reed2016generative,zhang2017stackgan,hong2018inferring}.}

\vspace{-0.7\baselineskip}

\paragraph{\textbf{Synthetic-to-Real Domain Adaptation.}} Previously, visual domain adaptation has been focused on the transformation of latent distributions in feature space \cite{sun2014from,long2015learning}. With the remarkably generative capability of GANs, new methods begin to convert synthetic images in pixel-level. Bousmalis {\it et al.} \cite{bousmalis2017unsupervised} combine a content-similarity loss with GANs to generate contexts around synthetic subjects mimicking the target domain. CyCADA \cite{hoffman2018cycada} combines cycle-consistency constraints with GANs to transfer the synthetic images at both pixel-level and feature-level. This line of work usually requires a dedicated loss formulation to balance the preservation of images contents. We avoid the loss design by using exactly content-preserving operations to generate color space variants of real images for training the classifier. RenderGAN \cite{sixt2017rendergan} inserts a parameterized 3D model with a cascade of image augmentations into the GANs to imitate the tag images rendering process. We focus on the color space adaptation of a synthetic dataset of diverse shapes for semantic segmentation, rather than rendering a template model from scratch for tag recognition.

\vspace{-0.7\baselineskip}

\paragraph{\textbf{Color Transformation.}} Learning color transformations, {\it e.g.,} enhancement, stylization and colorization, from given image exemplars is a challenging problem \cite{wang2011example,kuanar2019night,Kuanar2018deep,kuanar2019low}. \revised{There are mainly two types of automatic methods for color enhancement: example-based and learning-based. Example-based methods directly transfer the color of an example image to another one by optimization \cite{Reinhard2001color,huang2010example1,huang2010example2}. Learning-based color enhancement is another dominant stream which learns a mapping function from the source images to the target images with training data.

Learning-based color enhancement can be accomplished in two ways: by local transformations and by globally parameterized transformations. For local transformations, images with the desired pixel colors and styles are directly produced by a deep convolutional neural network \cite{yan2016automatic,gharbi2017deep,chen2018deep,limmer2016infrared}. These methods usually work best under a fixed image resolution and do not generalize well to images with arbitrary high resolutions, which lead to deteriorated results or unaffordable memory consumption. Recently, globally parameterized transformations have been proposed \cite{Park2018distort,hu2018exposure,Bianco2019learning,chai2020supervised}, in which the deep neural networks determine a series of parametric operations to enhance the input image. These methods can be further divided according to the usage of paired data \cite{Bianco2019learning} or unpaired data \cite{Park2018distort,hu2018exposure,chai2020supervised}, where the requirement of paired data often limits its usage. Hu {\it et al.} \cite{hu2018exposure} generates a retouching sequence using reinforcement learning (RL) with an adversarial reward. Park {\it et al.} \cite{Park2018distort} proposes a distort and recovery training scheme for unpaired learning using RL. While RL-based methods are powerful for learning complex behaviors, they are sensitive to the initial values and often requires a careful fine-tuning of the hyper-parameters to stabilize the training process. In contrast, our method does not rely on RL and is both robust and easy-to-use. Moreover, a direct application of these methods to our dataset would lead to poor performance due to a different kind of domain gap. The rendering images and the real images in our dataset does not share a common set of underlying scenes, which makes the differences in shapes and textures, rather than colors, easily dominate the classification. See the comparison in Section ~\ref{subsec:segmentation} for a more detailed discussion.

%

}

\vspace{-0.7\baselineskip}

\paragraph{\textbf{Image Segmentation.}} As a fundamental task in computer vision, pixel-level labeling ushers in a new development in semantic segmentation \cite{long2015fully,chen2018deeplab,bi2018dual,bi2017stacked,wang2017learning} and instance segmentation \cite{he2020mask}. Synthetic data have also been generated for training segmentation models \cite{richter2016playing}. We adapt the synthetic data to real-world images to bridge their domain gap in color space and consider this an effective way to improve segmentation performance.

\vspace{-0.5\baselineskip}

\section{METHOD}\label{sec:method}

\vspace{-0.5\baselineskip}

\begin{figure}[!t]
	\centering
	\includegraphics[width=\linewidth]{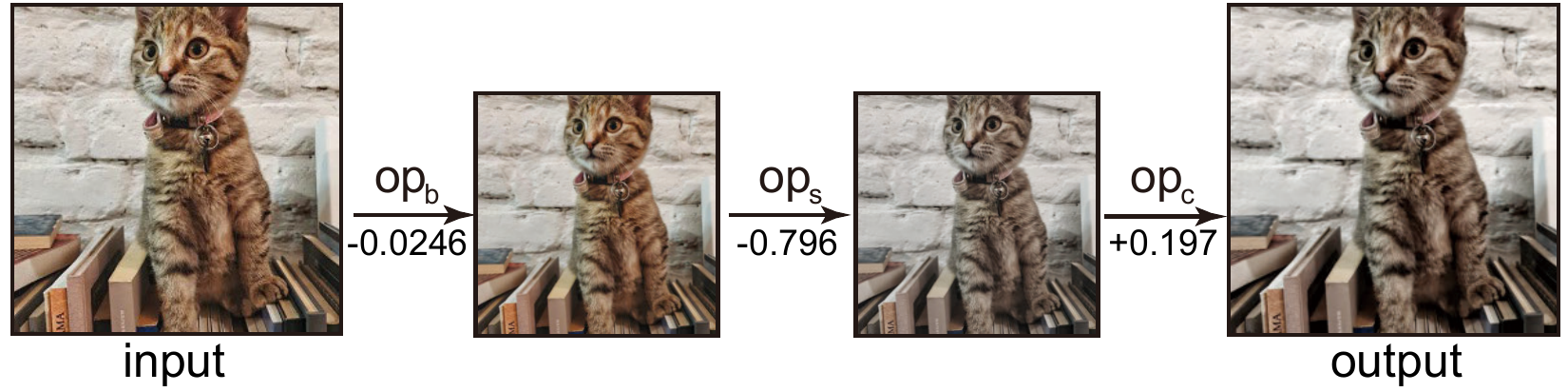}
	\caption{\textsc{Adjustment process.} A predicted sequence of color adjustment operations for transferring artist style A to artist style B on the Pexels dataset.}
	\label{fig:adjustment}
\end{figure}

The color space adaptation is a learning approach that is trained on unpaired images. \revised{Given the synthetic image dataset $X_s$ as source and the real image dataset $X_r$ as target, we expect to learn a function $\mathsf{g}_{s \rightarrow r}$ to adapt $X_s$ towards $X_r$ in color space, while preserving their textures and shapes. Fig.~\ref{fig:adap_results} shows the color space adaptation from synthetic cirrus cloud images to real images. }

\revised{To limit the adaptation only in color space, we transfer the images merely through a set of basic image processing operators, {\it i.e.,} adjusting the brightness, saturation, and contrast of the images, instead of through general convolutional kernels. We define the composition of the set of basic image processing operators as the adaptation operator $\mathsf{g}_{s \rightarrow r}(x; \alpha) := \mathsf{ops}(x; \alpha)$, where $x$ is the input image, $\alpha$ is the set of parameters for image processing, and $\mathsf{ops}$ is the combination of basic color adjustment operators. In our case,
\begin{equation}\label{eqn:gmin}
    \begin{aligned}
        \mathsf{g}_{s \rightarrow r}(x; \alpha) &= \mathsf{ops}(x; \alpha) \\
                                                &=\mathsf{op}_c(\mathsf{op}_s(\mathsf{op}_b(x; \alpha_{b}); \alpha_{s}); \alpha_{c}),
    \end{aligned}
\end{equation}
where $\alpha = \{\alpha_b, \alpha_s, \alpha_c\}$. Fig.~\ref{fig:adjustment} shows an example of the adjustment process with $\mathsf{ops}$. The closed-form expressions of $\mathsf{op}_b$, $\mathsf{op}_s$ and $\mathsf{op}_c$ can be found in Appendix~\ref{sec:appendix}.


Formally, we have an optimization problem defined as
\begin{equation}\label{eqn:gmin}
  \min_{\alpha} \sum_{x_{s} \in X_{s}} \sum_{x_{r} \in X_{r}} \mathsf{d}(\mathsf{g}_{s \rightarrow r}(x_{s}; \alpha), x_{r})
\end{equation}
where $x_{s}$ and $x_{r}$ is an image in $X_{s}$ and $X_{r}$, respectively, and $\mathsf{d}$ is a function measuring the distance between two images in their color space representations.

Here we pretrain a discriminator to measure the distance $\mathsf{d}$ and minimize a generator to produce adjustment parameters $\alpha$ for every image.
}

\vspace{-2\baselineskip}

\revised{
\textbf{\subsection{Generative Adversarial Network}\label{subsec:GAN}}

\vspace{-0.5\baselineskip}

GANs are proposed for estimating generative models via an adversarial process \cite{goodfellow2014generative}. In the general framework of GANs, there are two models trained alternatively: a generative model $G$ that captures the data distribution, and a discriminator model $D$ that estimates the probability that a sample comes from the training data rather than $G$. Both models are realized as convolutional neural networks in our case. The training procedure for $G$ is to maximize the probability of $D$ making a wrong prediction, while $D$ is trained to minimize that probability. As a result, $D$ and $G$ play the following two-player minimax game with value function $V(G, D)$:
\begin{equation}
    \begin{aligned}
      \min_{\theta_{g}} \max_{\theta_{d}} V(G, D) =
       \mathbb{E}_{x \sim p_{data}(x)} [\log D(x; \theta_{d})] \\ + \mathbb{E}_{z \sim p_{z}(z)} [\log (1 - D(G(z; \theta_{g}); \theta_{d}))]
    \end{aligned}
\end{equation}
where $x$ is training data and $z$ is noise. $\mathbb{E}$ denotes the expectation. Here we use the notion $x \sim p_{data}(x)$ for samples $x$ drawn from a corresponding probability data distribution $p_{data}$. The same also applies for $z \sim p_{z}(z)$. $G(z; \theta_{g})$ represents a function with parameters $\theta_{g}$ that maps from noise $z \sim p_{z}(z)$ to data space $x \sim p_{data}(x)$. For $G$ represented as convolutional neural networks, $\theta_{g}$ are parameters of convolutional kernels in the network. $D(x; \theta_{d})$ with parameters $\theta_{d}$ estimates the probability that $x$ came from the data rather than being produced by $G$. Again, $\theta_{d}$ are the parameters of a convolutional neural networks $D$. We will omit $x \sim p_{data}(x)$ and $z \sim p_{z}(z)$, as well as $\theta_{g}$ and $\theta_{d}$ for simplicity below.



\begin{figure}[!t]
	\centering
	\subfloat[]{\includegraphics[width=.33\linewidth]{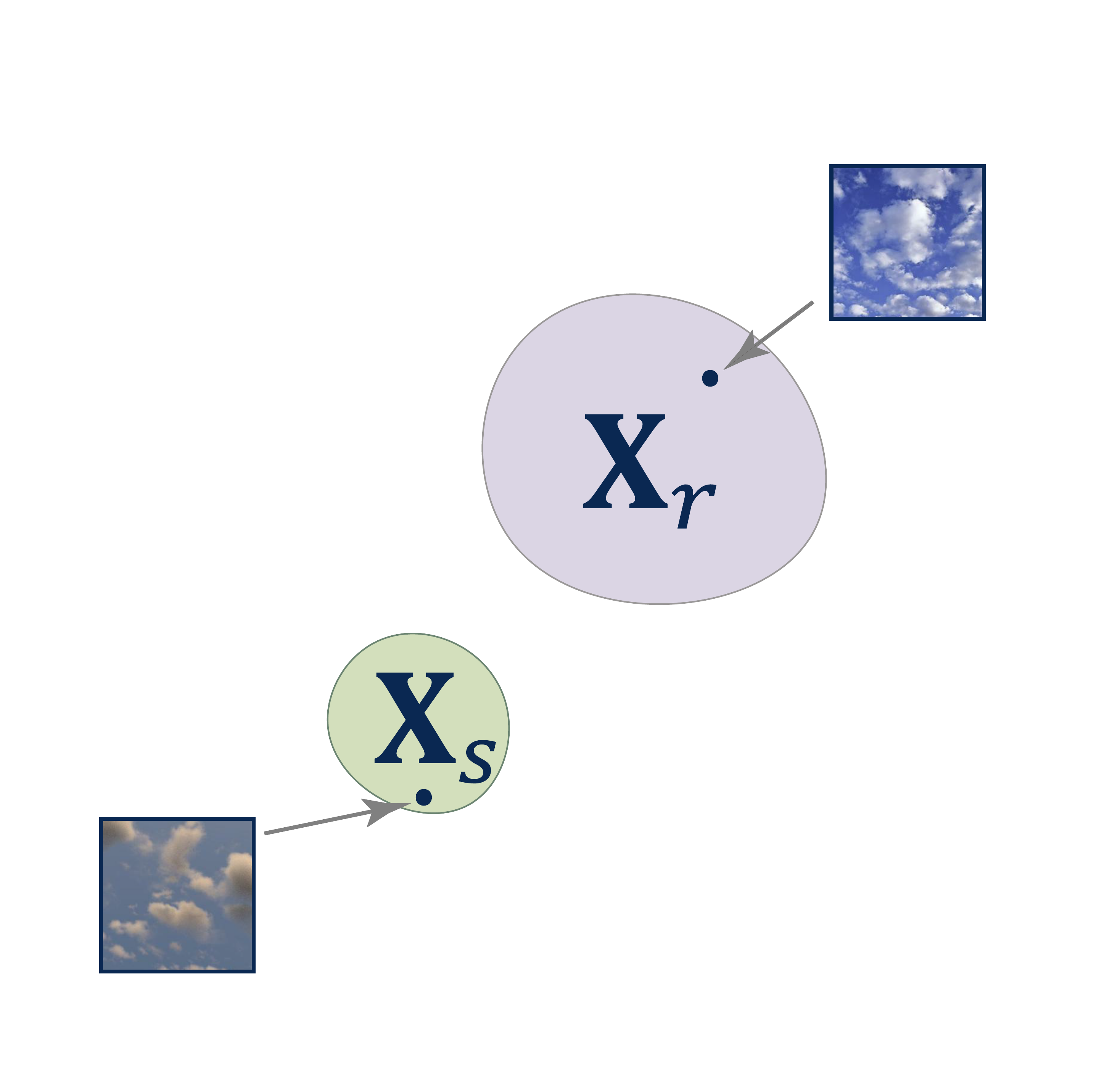}}
	\subfloat[]{\includegraphics[width=.33\linewidth]{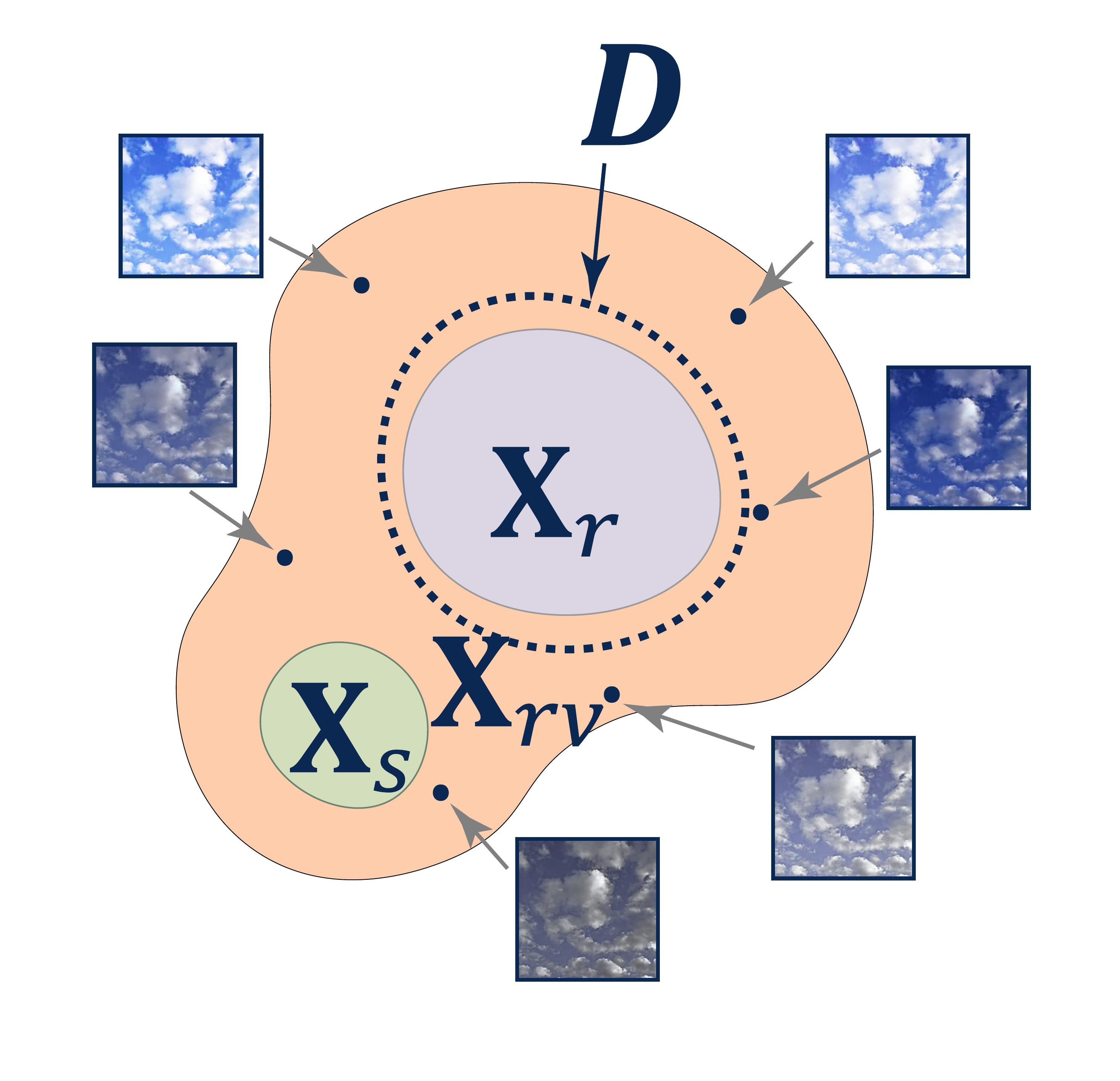}}
	\subfloat[]{\includegraphics[width=.33\linewidth]{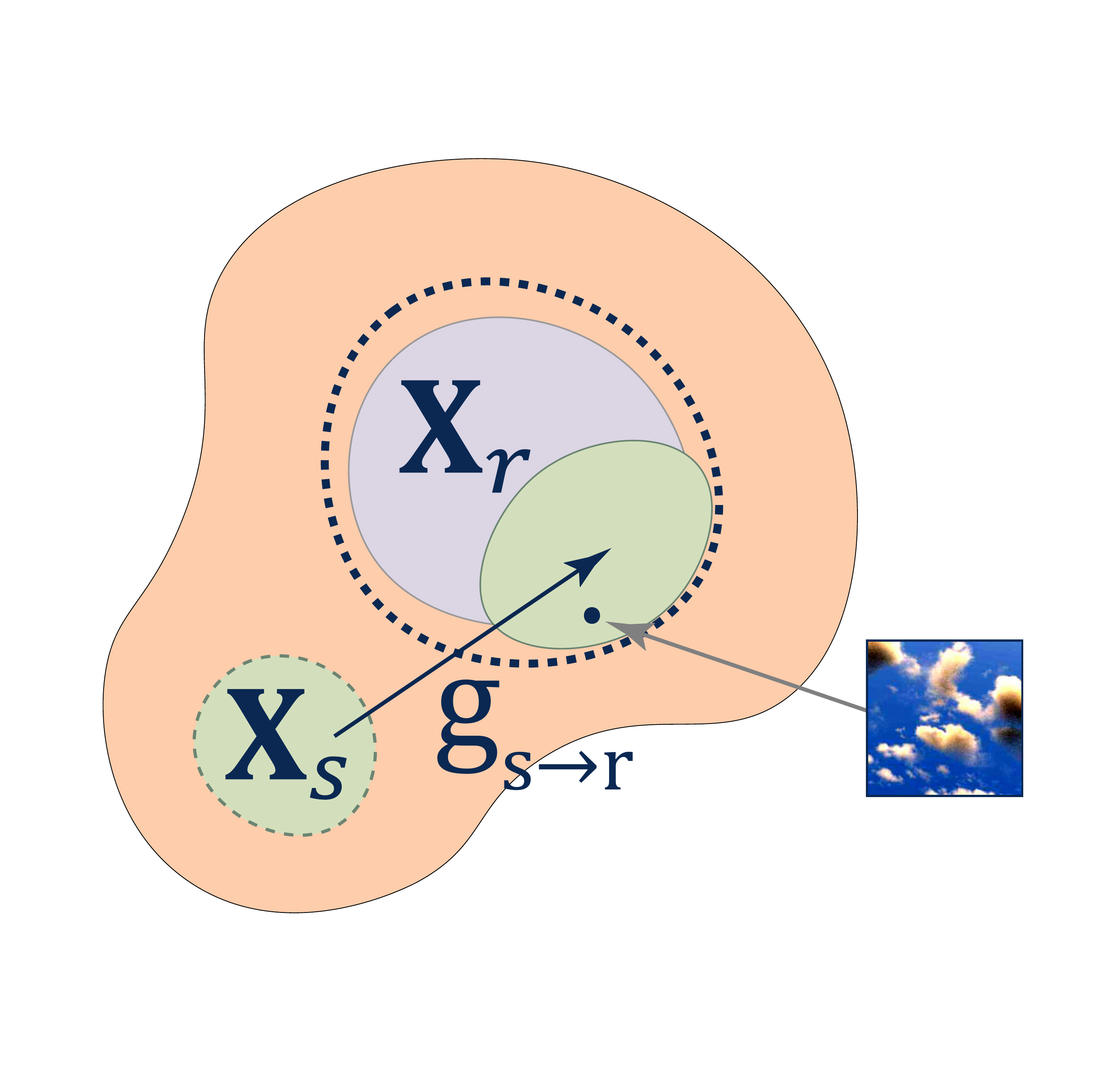}}\\
	\caption{\textsc{Method.} The synthetic images $X_s$ and real images represent distinct subsets $X_r$ in the color space (a). Color adjustment operations are used to generate random variants $X_{rv}$ of the real images $X_r$. A discriminator $D$ is trained to classify $X_{rv}$ and $X_r$ well in color space (b). Next, the generator is trained to adapt the synthetic images $X_s$ under the guidance of the pretrained discriminator $D$. The transferred synthetic images $\mathsf{g}_{s \rightarrow r}(X_s)$ after the adaptation is much more close to the real images $X_r$ in the color space (c).}
	\label{fig:high_level_design}
\end{figure}

\vspace{0.5\baselineskip}

\paragraph{\textbf{Color-Sensitivity.}} The standard GANs train the discriminator $D$ by classifying the transferred synthetic images $\mathsf{g}_{s \rightarrow r}(X_s)$ and the real images $X_r$. However, this quickly results in a color-insensitive discriminator due to the dominance of perceptual content differences, ({\it e.g.,} shapes, textures) between $X_s$ and $X_r$. We eliminate the undesired effects from content differences by purposely constructing a training dataset whose images are different only in the color space. Specifically, we generate the color variants of real images $X_r$ as $X_{rv}$ by using the image processing operators $\mathsf{ops}$. When training the discriminator $D$, color variants $X_{rv}$ are seen as \textit{False} and the original real images $X_r$ are seen as \textit{True}. By classifying $X_{rv}$ and $X_r$, we succeeds to learn a color space discriminator $D$ with its classification boundary found in color space (Fig.~\ref{fig:high_level_design}-(b)). While we prevent $D$ from directly observing the synthetic data $X_s$ during training, the color space discriminator $D$ still works for synthetic data $X_s$ since the color variants $X_{rv}$ have covered color space of $X_s$ (Fig.~\ref{fig:high_level_design}-(b)). Under the guidance of the pre-trained $D$, we can learn the generator $G$ to adapt the color space information of $X_s$ correctly (Fig.~\ref{fig:high_level_design}-(c)).
}

\begin{figure}[!t]
	\centering
	\includegraphics[width=\linewidth]{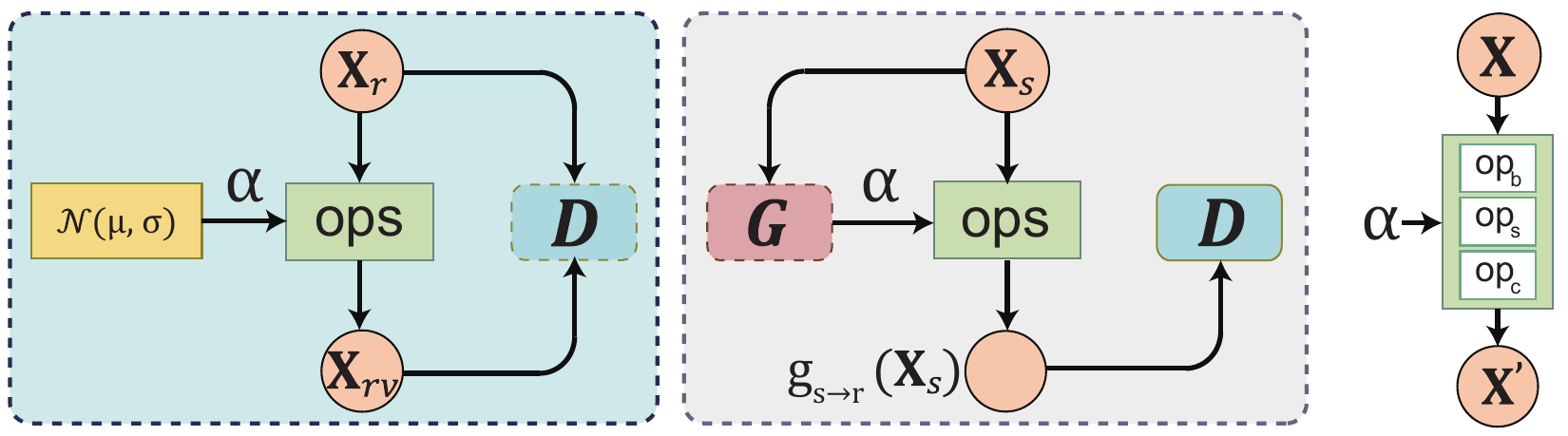}
	\caption{\textsc{Pipeline.} We learn the discriminator $D$ (left) by pretraining on $X_r$ and $X_{rv}$ and then train the generator $G$ (middle) under the guidance of the pretrained $D$. The module of color adjustment operations $\mathsf{ops}$ (right) is shared between the two stages. It takes in Gaussian random samples as parameters to generate adversarial data for training the discriminator, and takes in the generator output as parameters to generate adapted synthetic images for training the generator.}
	\label{fig:pipeline}
\end{figure}

\vspace{-2\baselineskip}

\revised{\textbf{\subsection{Color Space Discriminator}}\label{subsec:color_space_d}

\vspace{-0.5\baselineskip}

As shown in Fig.~\ref{fig:pipeline}-left, }to embody the discriminative function $\mathsf{d}$, we choose to sample sufficient variants of $X_r$ in color space as $X_{rv}$, and consequently train a classifier to distinguish between $X_r$ and the adversarial dataset $X_{rv}$. Therefore, we adopt the composition of a set of basic image processing operators $\mathsf{ops}$ \revised{(Fig.~\ref{fig:pipeline}-right)}, to randomly adjust the brightness, saturation and contrast of images in $X_r$. These low-level operators ensure that the learned classifier is only effective in distinguishing the color representations of the images, but insensitive to their high-level perceptual difference. \revised{We use $D$ as the discriminative function $\mathsf{d}(x_{rv},x_r)$ and determine its parameters by optimizing the loss function:
\begin{equation}\label{eqn:dmin}
    \mathit{L}_D = \mathbb{E} \left[-\log (D(x_r))\right] + \mathbb{E} \left[-\log\left(1-D(x_{rv})\right)\right]
\end{equation}
}where $x_{rv} = \mathsf{ops}(x_r; \alpha_{\mathsf{randn}})$ is an image in dataset $X_{rv}$ and $\alpha_{\mathsf{randn}} \sim \mathcal{N}(\mu,\sigma)$. We compute the mean $\mu$ and standard deviation $\sigma$ of $\mathcal{N}$ as
\begin{equation}\label{eqn:ndist}
  \mu = 0, \quad \int_{-1}^{1}\frac{1}{\sqrt{2\pi\sigma^{2}}}e^{\frac{-(\alpha-\mu)^{2}}{2\sigma^{2}}}=p
\end{equation}
where \(p\) is the probability that $\alpha$ falls in the range $[-1, 1]$. We set $p = 0.99$ to ensure that $\alpha$ is in the usual range of color adjustment in most cases.

\vspace{-1\baselineskip}

\revised{\textbf{\subsection{Color Adjustment Parameter Generator}}\label{subsec:color_adjust_param_g}

\vspace{-0.5\baselineskip}

As shown in Fig.~\ref{fig:pipeline}-middle, }the color adjustment operations $\mathsf{ops}$ are embedded into $\mathsf{g}_{s \rightarrow r}$, by taking the output of the generator $G$ as the adjustment parameters $\alpha$. In some sense, this mimics the post-processing steps taken by photographers in the real world to retouch raw images. Specifically, the generator is composed of three components with the same network structure, {\it i.e.} $G = \{G_b,G_s,G_c\}$. Each component takes an input image and produces an adjustment parameter for an operator in $\mathsf{ops}$ respectively, $\alpha = \{\alpha_b=G_b(x),\alpha_s=G_s(x),\alpha_c=G_c(x)\}$. Finally, the $\mathsf{ops}$ together with the produced $\alpha$ are applied on the input image to generate an adapted one. \revised{
We train the generator $G$ by maximizing the score of the adapted images $\mathsf{g}_{s \rightarrow r}(X_s)$ obtained on the pre-trained discriminator $D$ (fixed in this stage). The loss function is:
\begin{equation}\label{eqn:gmin}
    \mathit{L}_G = \mathbb{E}[\log (1 - D(\mathsf{g}_{s \rightarrow r}(x_{s}; G(x_s))))]
\end{equation}
}

\vspace{-4\baselineskip}

\revised{\textbf{\subsection{Training}}

\vspace{-0.7\baselineskip}

Our training process has two stages. In the first stage (Fig.~\ref{fig:pipeline}-left), we feed the adversarial data $X_{rv}$ and real images $X_r$ into the $D$ to train its parameters by following Eqn.~\ref{eqn:dmin}. In the second stage (Fig.~\ref{fig:pipeline}-middle), we fix the pre-trained $D$ and feed the synthetic images $X_s$ into the $G$ to train its parameters by following Eqn.~\ref{eqn:gmin}. However, with our color-sensitive discriminator, we need not follow the standard GANs training process to repeat the two-stages alternatively, which simplifies the GAN training and makes our method robust.
}

\vspace{1.5\baselineskip}

\revised{\section{ARCHITECTURE AND TRAINING DATA}}
\vspace{-2\baselineskip}

\revised{

\textbf{\subsection{Datasets}}\label{sec:synthetic_data}

\vspace{-0.5\baselineskip}

\paragraph{\textbf{The SynCloud Dataset:}} We construct the SynCloud dataset for quantitative evaluation on the semantic segmentation task of the cirrus cloud images. It contains synthetic cloud images naturally attached with semantic segmentation labels (tagged as {\it Synthetic Part}), and real cloud images (tagged as {\it Real Part}).
\begin{enumerate}[-]
	\item {{\it Synthetic Part.} We construct $11,654$ synthetic images, $7,969$ images of which for training, and the left $3,685$ images for testing, based on the shape modeling and photo-realistic 3D rendering methods. First, we computationally generate $624$ 3D volume data with heterogeneous grid densities, through fluid simulation and image-based reconstruction. Then, we use a physically-based renderer Mitsuba \cite{jakob2010mitsuba}, to generate photo-realistic images, as well as the segmentation labels, from those volume data of cirrus clouds. We randomly sample the parameters of the volume rendering, {\it e.g.,} scene lighting, reflections, lens focal length, and camera viewpoint, to extend the coverage range. The generated synthetic images all have $512\times340$ resolution. \footnote{We will release the cirrus clouds dataset, including all the volume data, rendering settings and rendering results.} }

    \item { {\it Real Part.} This part of the data contains $1,381$ unlabeled real images for adaptation training and $427$ real images with manually annotated labels for segmentation training. Among annotated images, $273$ of them are in the training set, and the left $154$ images are in the test set. We manually annotate the images using the Magnetic Lasso tool in Adobe Photoshop. Due to the complex boundaries of cirrus, manual labeling is exceptionally time-consuming. It takes around 10-20 minutes per image. }
\end{enumerate}

\vspace{-1.\baselineskip}

\paragraph{\textbf{Photo Datasets for Style Transfer:}} For the style transfer and photo post-processing task, we utilize two sources of training data:
\begin{enumerate}[-]
	\item {{\it The MIT-Adobe FiveK Dataset.}  It is a photo dataset consisting of $5,000$ RAW images and their retouched versions created by five professional experts \cite{bychkovsky2011learning}. In this work, we choose $4,000$ input RAW images and $4,000$ retouched images of expert A as training data and use another $1,000$ input RAW images as test data. }
    \item {{\it The Pexels Dataset.}  We crawled professionally retouched photos from two artists on \emph{pexels.com}. The dataset consists of $284$ images from artist A and $425$ images from artist B. We use $90\%$ of them for training and $10\%$ for testing. The two sets of data have relatively consistent styles. One of them is bright and vivid, and the other is more cold and tranquil. We show some exemplars in Fig.~\ref{fig:dataset}.}
\end{enumerate}
All images in the MIT-Adobe FiveK dataset and the Pexels dataset are resized with a fixed width of $256$ and randomly cropped to $256\times256$ resolution.

}

\vspace{-2\baselineskip}

\begin{figure}[!t]
	\centering
	\includegraphics[width=\linewidth]{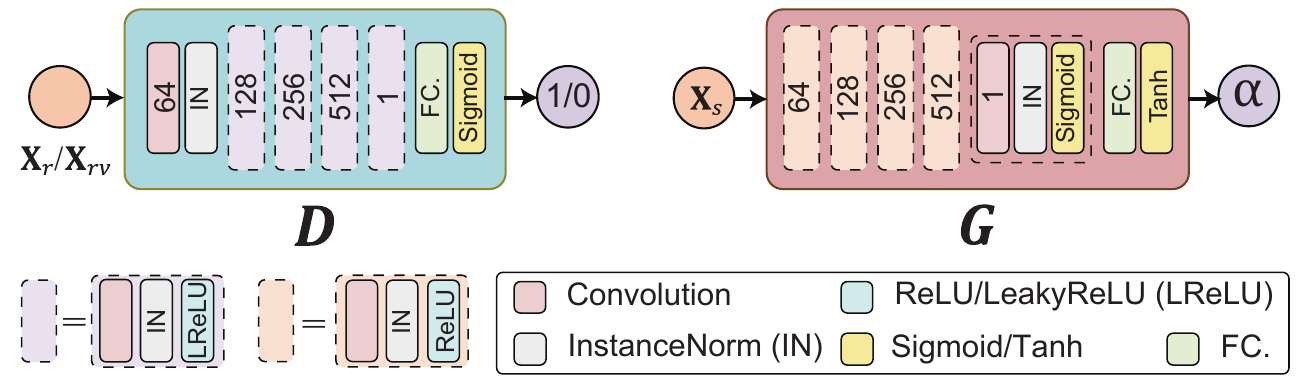}
	\caption{\textsc{Architecture.} The network architectures of the $D$ and $G$. \revised{All convolution layers have filters of kernel size $4$ and strides $2$. The number of output filters is written in the block.} }
	\label{fig:architecture}
\end{figure}

\revised{\textbf{\subsection{Network Architecture}}

\vspace{-0.5\baselineskip}

\paragraph{\textbf{Discriminator}}
We design the network structure of the discriminator $D$ (Fig.~\ref{fig:architecture}-left) by following the guidelines of DCGANs \cite{radford2016unsupervised}. Specifically, we use strided convolutions to replace pooling layers and choose Leaky ReLUs (with slope coefficient $0.2$) as the activation function. We also use instance normalization instead of batch normalization. The convolution group, including a convolution layer, an instance normalization layer, and a leaky ReLU layer, are repeated four times. At last, a fully connected layer is connected with a sigmoid function as the output.

\vspace{-1\baselineskip}

\paragraph{\textbf{Generator}}
We design the network structure of the generator $G$ (Fig.~\ref{fig:architecture}-right) by following the guidelines of DCGANs \cite{radford2016unsupervised}. The convolution group, including a convolution layer, an instance normalization layer, and an activation function layer, are repeated five times. At last, a fully connected layer is connected with a tanh function as the output.
}

\vspace{-2\baselineskip}

\revised{\textbf{\subsection{Training Details}}}

\vspace{-0.5\baselineskip}

\begin{figure}[!t]
	\centering
	\subfloat[]{\includegraphics[width=.5\linewidth]{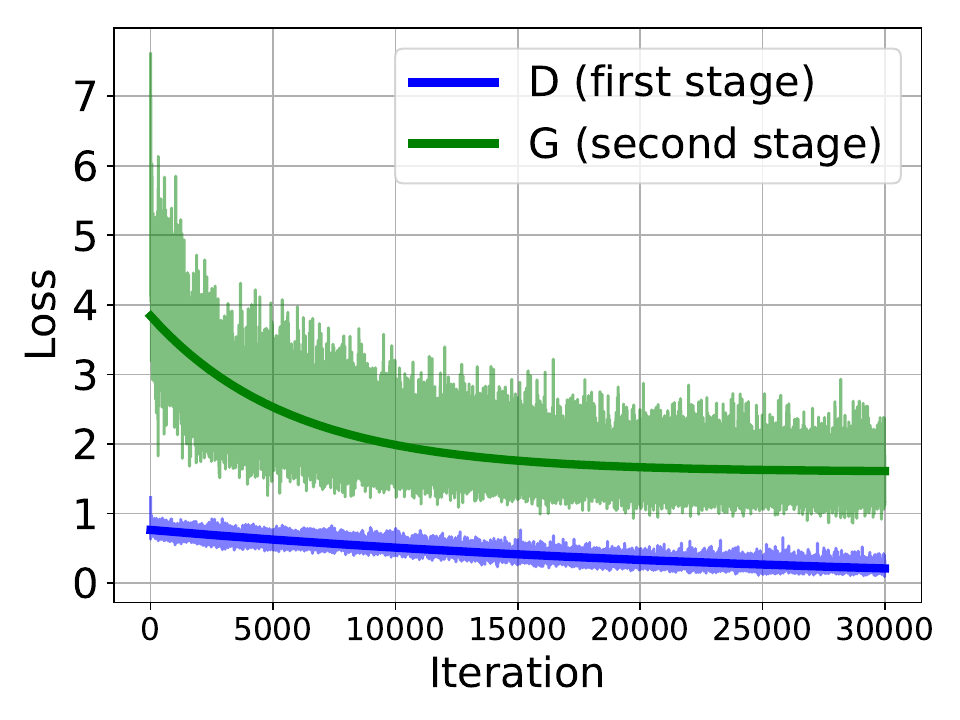}}
	\subfloat[]{\includegraphics[width=.5\linewidth]{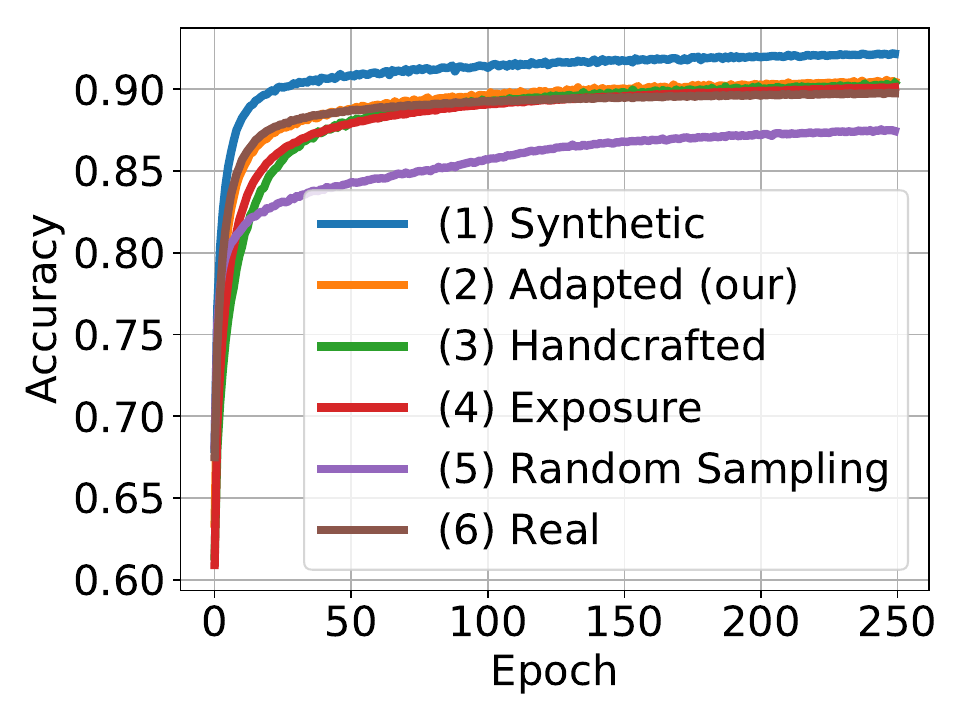}}\\
	\caption{\revised{\textsc{Training Process.} (a) Losses of the discriminator and the generator, during the first and second stage of the color space adaptation training. (b) The FCN accuracies during the semantic segmentation training. }}
	\label{fig:training_curve}
\end{figure}

All our implementations are based on the Keras library \cite{chollet2015keras} with the TensorFlow \cite{abadi2016tensorflow} backend. A desktop machine with Ubuntu 16.04 and a Geforce 1080Ti GPU is used for training the neural networks.

\vspace{-1\baselineskip}

\paragraph{\textbf{Adaptation Training.}}
For all the experiments, we choose $24$ as the batch size. The input images are zero-centered and rescaled to $[-1, 1]$. The network weights are initialized with He initialization \cite{he2015delving}. We adopt the Adam optimizer \cite{kingma2015adam}, and train all the networks from scratch with $lr=1\mathrm{e}{-3}$ and $\beta_1=0.5$. First, we train the discriminator for \revised{$4$ epochs}. Then the generator is trained for \revised{$4$ epochs} with the pretrained discriminator frozen. \revised{The loss change during the training process is shown in Fig.~\ref{fig:training_curve}-(a).} When training $D$, we get a loss value around $0.32$ for training data and $0.25$ for test data. When training $G$, we get a loss value of around $1.39$ for training data and $1.83$ for test data.

Partial results of the synthetic images before and after color space adaptation are shown in Fig.~\ref{fig:adap_results}. Visually, the adapted images are much more similar to the color style of real images. \revised{The sliced-Wasserstein distance}\footnote{ \revised{SWD owns similar properties to the Wasserstein distance but simpler to compute. It is widely used in various applications, including generative modeling and general supervised/unsupervised learning, to measure the quality of generative images \cite{Kolouri2019generalized}.}} (SWD) metric \cite{bonneel2015sliced} between the synthetic images and the real images has reduced about $55\%$ after the adaptation. 

\vspace{-1\baselineskip}

\paragraph{\textbf{Segmentation Training.}}
The FCN-8s architecture \cite{long2015fully} is used to perform the semantic segmentation task. We fine-tune network with the VGG-16 model and adopt the SGD optimizer with the learning rate of $1\mathrm{e}{-5}$ and the momentum of $0.9$. Mean squared error is used as the loss function. We iterate the training process for $250$ epochs and achieve a stable training accuracy of around $0.9$. \revised{The accuracy change during the training process is shown in Fig.~\ref{fig:training_curve}-(b).}

\section{EXPERIMENTS}\vspace{-0.5\baselineskip}

\vspace{-1.5\baselineskip}

\textbf{\subsection{Evaluation}} \label{subsec:segmentation}

\vspace{-0.5\baselineskip}

\paragraph{\textbf{Baseline and Comparison.}}We quantitatively evaluate the color space adaptation method through the performance of semantic segmentation tasks. Specifically, we prepare six datasets for training the segmentation network:
\begin{enumerate}[(1)]
	\item {{\it Synthetic images \revised{(from {\it Synthetic Part})}.} $7,969$ synthetic images with automatic generated segmentation labels through physically-based rendering.}
	\item {{\it Adapted synthetic images \revised{(our)}.} \revised{$7,969$ synthetic images adapted with our method.}}
    \item {{\it Augmented synthetic images by a handcrafted approach.} $7,969$ synthetic images augmented by fitting a Gaussian model and shifting the features. We show details in Appendix~\ref{sec:handcrafted approach}.}
    \revised{\item {{\it Enhanced synthetic images by Exposure \cite{hu2018exposure}.} $7,969$ synthetic images retouched by the method of Exposure.}}
    \item {{\it Augmented synthetic images by random sampling.} $21,252$ augmented images enhanced from $7,969$ synthetic images by randomly sampling the color adjustment operation parameters.}
    \item {{\it Real images \revised{(from {\it Real Part})}.} $273$ real images with manually annotated segmentation labels.}
\end{enumerate}

Finally, we use $154$ unique test real images \revised{(from {\it Real Part})} to evaluate the segmentation performance of the FCNs trained on the above six datasets. \revised{We use mean IoU scores to measure the segmentation performance. Mean IoU is the intersection over the union of the prediction and ground truth in the segmentation results. A higher score means better performance.}

Table~\ref{tab:basic_compare} shows the evaluation results. Compared with the original synthetic images, the randomly augmented synthetic images has only less than $2\%$ relative improvement. On the other hand, using the adapted synthetic images as training data has improved the segmentation performance from $0.68$ to $0.75$ (about $10\%$ relative improvement), proving the efficacy of our color space adaptation method. The performance of adapted images is very close to that of the real images. It demonstrates that replacing real images with synthetic images to eliminate manual labeling efforts is promising, at least partially. Using the combined dataset with the adapted images and real images, we obtain a segmentation performance of $0.7765$, which is higher than merely using the real images.

\revised{We also execute a comparison with a state-of-the-art method for color enhancement \cite{hu2018exposure}. Table~\ref{tab:basic_compare} shows that directly applying the method to our dataset does not perform well on the segmentation task. The method also adopts an adversarial loss based on a discriminator to guide the inference learning of the color adjustment process. Their source and target images for training the discriminator share a common set of underlying scenes, which naturally form a color-sensitive dataset and result in a color-sensitive discriminator. However, in our dataset the synthetic and real images have disjoint sets of underlying scenes, whose differences in shapes and textures, but not colors, would easily dominate the classification. This inherent gap in our synthetic-to-real dataset together with the complexity of the reinforcement learning makes it challenging to obtain a color-sensitive discriminator, which explains the failure. The segmentation performance of $0.6176$ is close to our ablation study result of $0.6247$ with a color-insensitive discriminator. This also justifies the necessity of our decoupled construction of the color variants $X_{rv}$ for training the discriminator.}

\setlength{\tabcolsep}{8pt}
\begin{table}[!t]
	\begin{center}
		\caption{mean IoU for all datasets.}
		\label{tab:basic_compare}
		\begin{tabular}{ll}
			\hline\noalign{\smallskip}
			Training Dataset $\qquad\qquad$ & mean IoU (test) \\
			\noalign{\smallskip}
			\hline
			\noalign{\smallskip}
            (1) Synthetic & 0.6829\\
            (2) Adapted (our) & \textbf{0.7488}\\
            (3) Handcrafted & 0.7118\\
            (4) Exposure \cite{hu2018exposure} & 0.6176\\
			(5) Random Sampling & 0.6996\\
			(6) Real & \textbf{0.7668}\\
			\hline
		\end{tabular}
	\end{center}
\end{table}
\setlength{\tabcolsep}{1.4pt}

\setlength{\tabcolsep}{6pt}
\begin{table}[!t]
	\begin{center}
		\caption{mean IoU for different adaptation options.}
		\label{tab:ablations}
		\begin{tabular}{lllllll}
			\hline\noalign{\smallskip}
			\multicolumn{3}{c}{Choice of $p$ in Eqn.~\ref{eqn:ndist}} &\multicolumn{2}{c}{Color-Sensitive $D$}&\multirow{2}*{mean IoU}\\
			0.9 & 0.99 & 0.999 & with & without\\
			\hline
			\checkmark & & & \checkmark & & 0.7359\\
			& \checkmark & & \checkmark & & \textbf{0.7488}\\
			& & \checkmark& \checkmark & & 0.7458\\
			\hline
			& \checkmark & & & \checkmark & 0.6247\\
			& & \checkmark & & \checkmark & 0.6753\\
			\hline
		\end{tabular}
	\end{center}
\end{table}
\setlength{\tabcolsep}{1.4pt}


\vspace{-0.5\baselineskip}

\paragraph{\textbf{Ablations.}}
\revised{We adopt the following two ablations:}
\begin{enumerate}[(1)]
	\item {\revised{{\it Choice of $p$ in Eqn.~\ref{eqn:ndist}.}} \revised{We choose different $p$ of $0.9$, $0.99$, and $0.999$ to control the probability of the adjustment parameter $\alpha$ in the range $[-1,1]$. Segmentation performance in Table~\ref{tab:ablations} (the first three lines)} shows that the change of $p$ brings slight variations to the mean IoU scores. In practice, we have found that $p=0.99$ is a reasonably good choice for all the experiments. }


    \item {\revised{{\it With/without the color-sensitive $D$.}} \revised{We train a color-sensitive $D$ by following the design in Section~\ref{subsec:GAN}, and a color-insensitive one by classifying between mapped source images and target images and putting it into the back-propagation. Table~\ref{tab:ablations} (the last two lines) shows that with a color-insensitive $D$}, the segmentation performance quickly degrades, to be even worse than using non-adapted images. As discussed in Section~\ref{subsec:GAN}, a possible reason is that once the discriminator starts to observe the synthetic images, the perceptual information like shapes and textures could gradually dominate the classification.}


\end{enumerate}


\vspace{-2\baselineskip}

\textbf{\subsection{Applications}}

\vspace{-0.5\baselineskip}

\begin{figure}[!t]
	\centering
	\includegraphics[width=\linewidth]{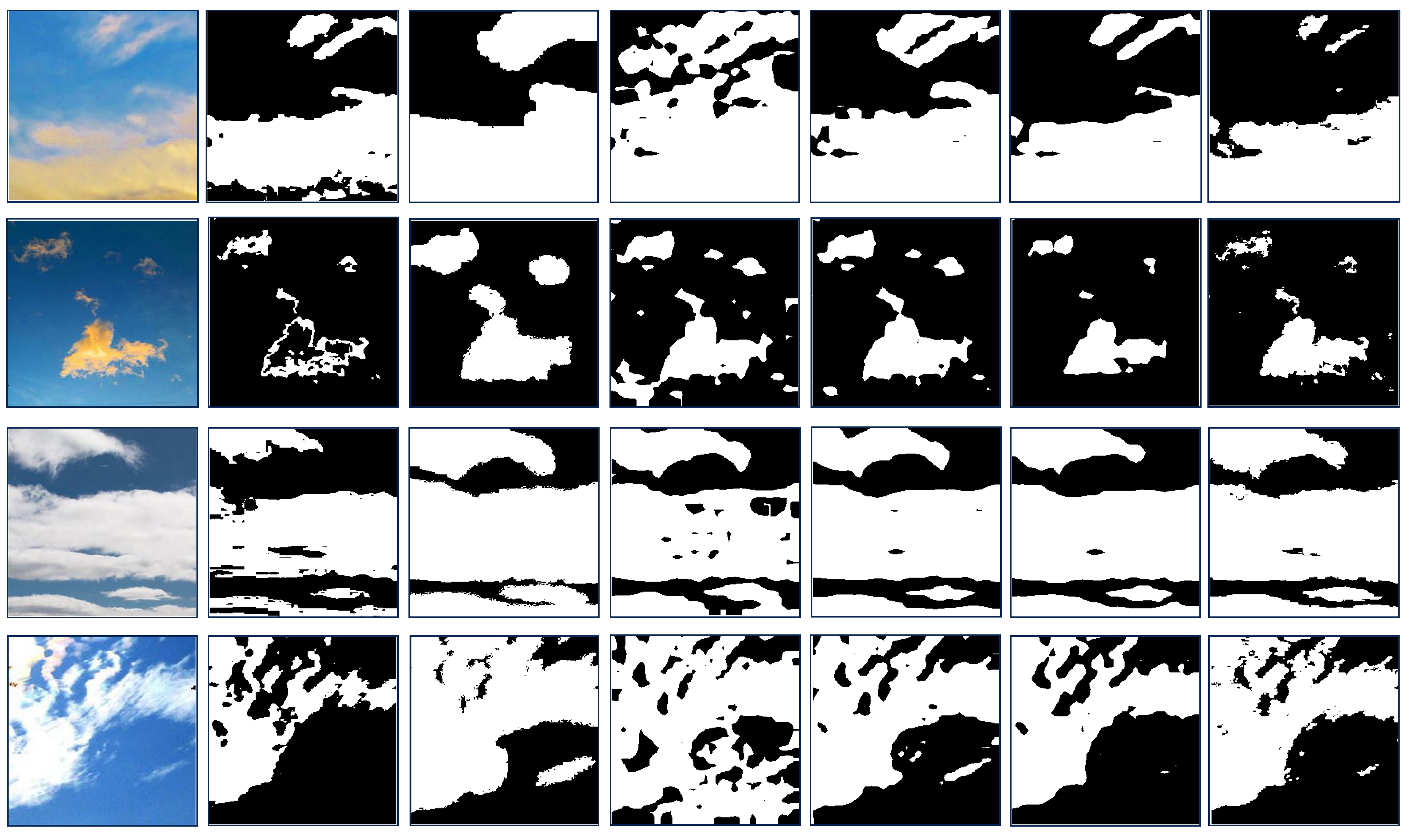} \\
	(a) \hspace*{7mm} (b) \hspace*{7mm} (c) \hspace*{7mm} (d) \hspace*{7mm} (e) \hspace*{7mm} (f) \hspace*{7mm} (g)\\
	\caption{\textsc{Segmentation Results.} (a) The original images. (b) The chromatic segmentation \cite{dobashi2010modeling}. (c) The paint selection segmentation \cite{liu2009paint}. (d) The segmentation trained on the original synthetic images. (e) The segmentation trained on the synthetic images after color space adaptation. (f) The segmentation trained on the manual labeled real images. (g) The ground truth segmentation labels.}
	\label{fig:seg_results}
\end{figure}

\paragraph{\textbf{Cirrus Segmentation.}}We present the method mainly for the image segmentation of cirrus clouds. Fig.~\ref{fig:seg_results} compares some segmentation results of different methods. The interactive segmentation methods (Fig.~\ref{fig:seg_results}-(b) and \ref{fig:seg_results}-(c)) are often tedious and time-consuming to yield a reasonable annotation result. The segmentation results using the original synthetic images for training have obvious artifacts (Fig.~\ref{fig:seg_results}-(d)). On the other hand, using the adapted synthetic images for training significantly improves the segmentation accuracy (Fig.~\ref{fig:seg_results}-(e)), whose results are visually quite similar with that of using manually labeled images for training (Fig.~\ref{fig:seg_results}-(f)) and the ground truth (Fig.~\ref{fig:seg_results}-(g)). Finally, we can make some further applications based on the segmentation results.

\vspace{-0.8\baselineskip}

\begin{figure}[!t]
	\centering
	\includegraphics[width=\linewidth]{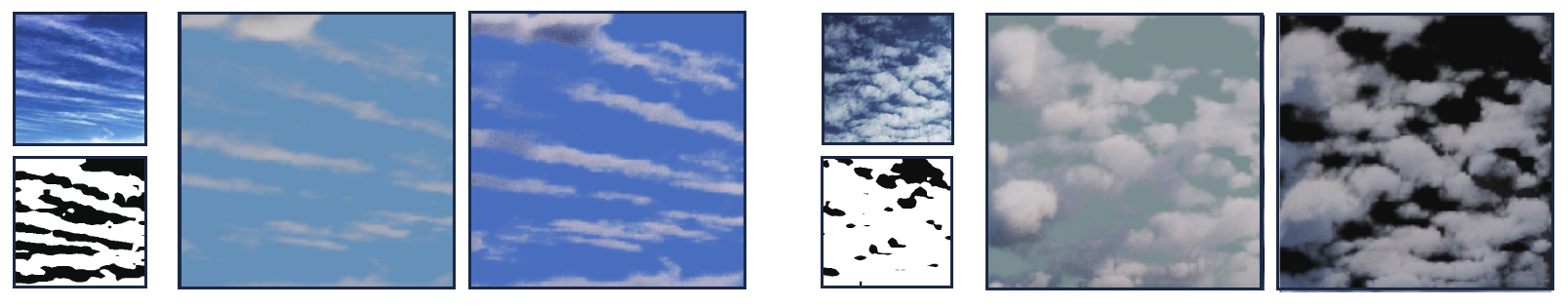}
	\caption{\textsc{Further Applications.} The single image based 3D volume data reconstruction of cirrus clouds.}
	\label{fig:app_recon}
\end{figure}

\paragraph{\textbf{3D Reconstruction.}}By leveraging the trained segmentation network, we generate the segmentation label as input to reconstruct the 3D volume data from a cirrus cloud image \cite{dobashi2010modeling}. We re-render it under new lighting conditions to produce new images (Fig.~\ref{fig:app_recon}).

\vspace{-0.8\baselineskip}

\begin{figure}[!t]
	\centering
	\includegraphics[width=\linewidth]{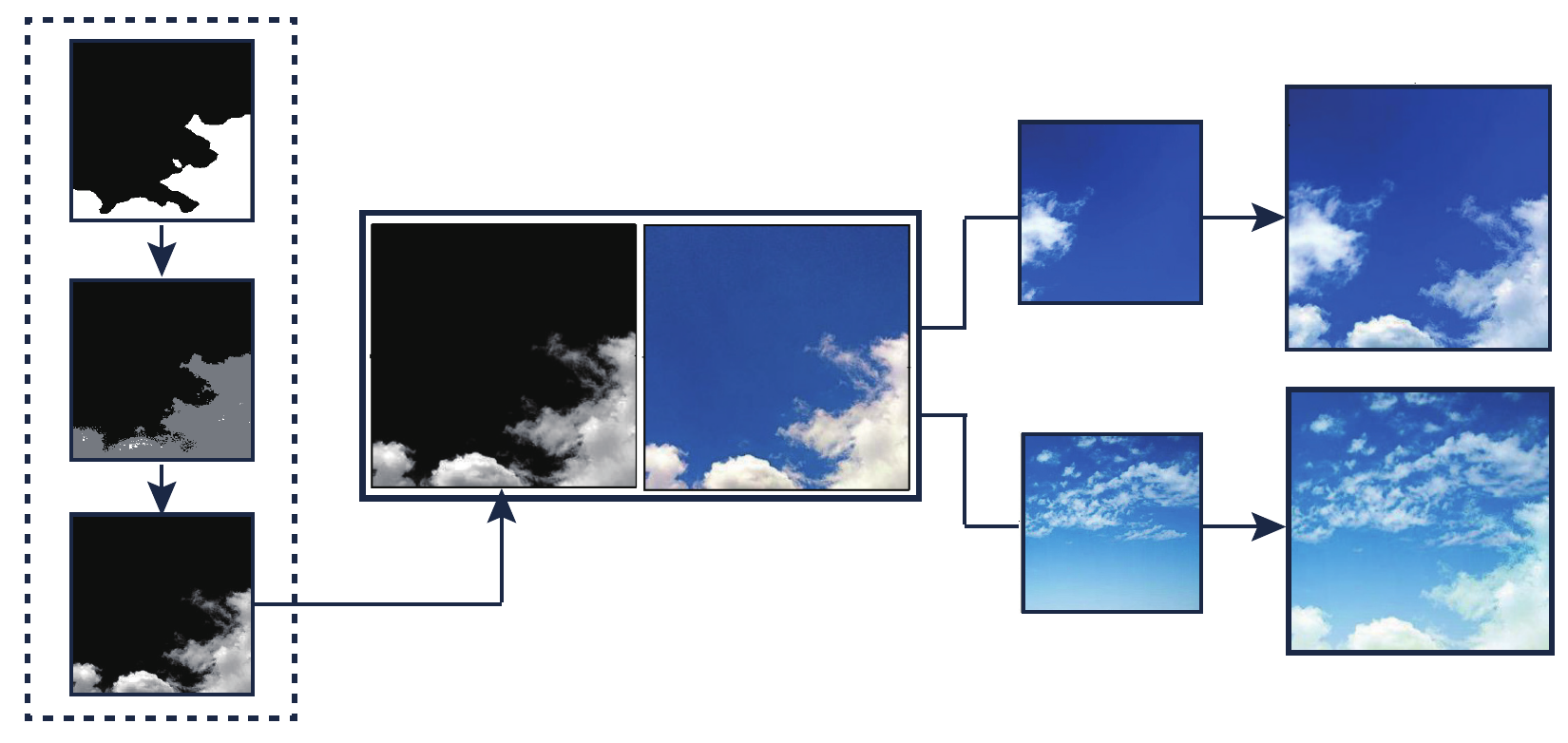}
	\caption{\textsc{Further Applications.} The cirrus clouds image matting and composition. }
	\label{fig:app_matt}
\end{figure}

\paragraph{\textbf{Matting and Composition.}}We execute image matting \cite{chen2013knn} based on a tri-map computed from the inflated segmentation label, to first separate the cirrus cloud foreground from the source image (Fig.~\ref{fig:app_matt}-left), and then paste it onto other target images (Fig.~\ref{fig:app_matt}-middle) to recompose novel cirrus clouds images (Fig.~\ref{fig:app_matt}-right).

\vspace{-0.8\baselineskip}

\begin{figure}[!t]
	\centering
	\includegraphics[width=\linewidth]{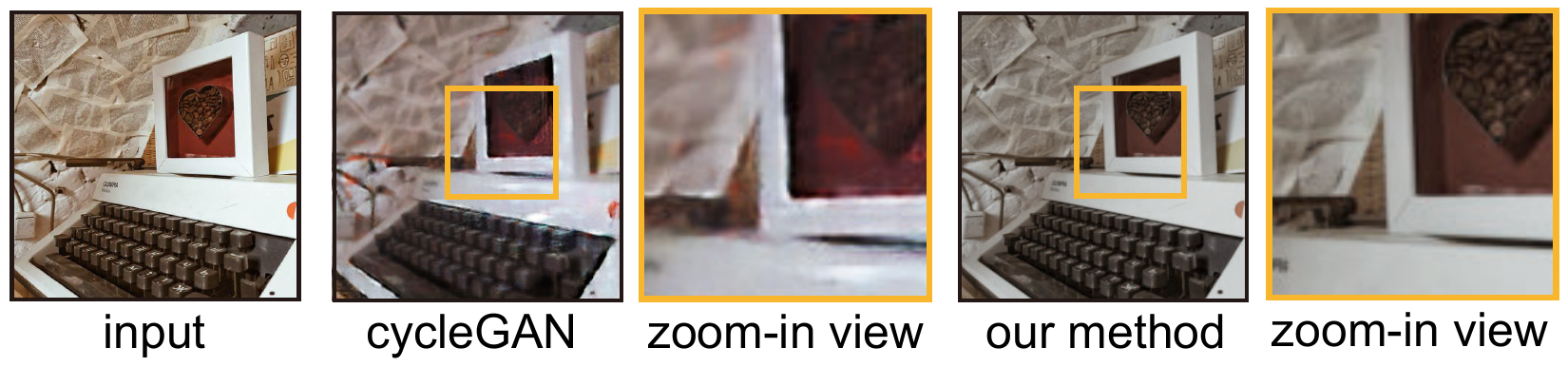}
	\caption{\textsc{Comparison with CycleGAN.} Our method can process arbitrary high-resolution images on which CycleGAN leads to blurry results. }
	\label{fig:compare_with_cycleGAN_pexels}
\end{figure}

\paragraph{\textbf{Style Transfer and Photo Post-Processing.}}Our method has the potential for tasks like style transfer and photo post-processing.
We execute style transfer task on the MIT-Adobe \cite{bychkovsky2011learning} dataset and the Pexels dataset, and compare our method with CycleGAN. We put experiment details in Appendix~\ref{sec:appendix_dataset}. Fig.~\ref{fig:compare_with_cycleGAN_fiveK} shows the transferred styles. Our method can output results with comparable quality to CycleGAN on these datasets. While CycleGAN generates images directly and owns a broader space to change styles, our method predicts parameters for human-understandable operations. Fig.~\ref{fig:adjustment} demonstrates a sequence of color adjustment operations for transferring between two artists' styles.
Moreover, CycleGAN leads to blurry images and edge distortion (see Fig.~\ref{fig:compare_with_cycleGAN_pexels}), while our method is inherently resolution-independent due to the content preserving operations. Therefore, we can train our model on low-res images, and apply it to arbitrary high-res ones, making the method more suitable for photo processing and graphics applications.


\begin{figure}[!t]
	\centering
	\includegraphics[width=\linewidth]{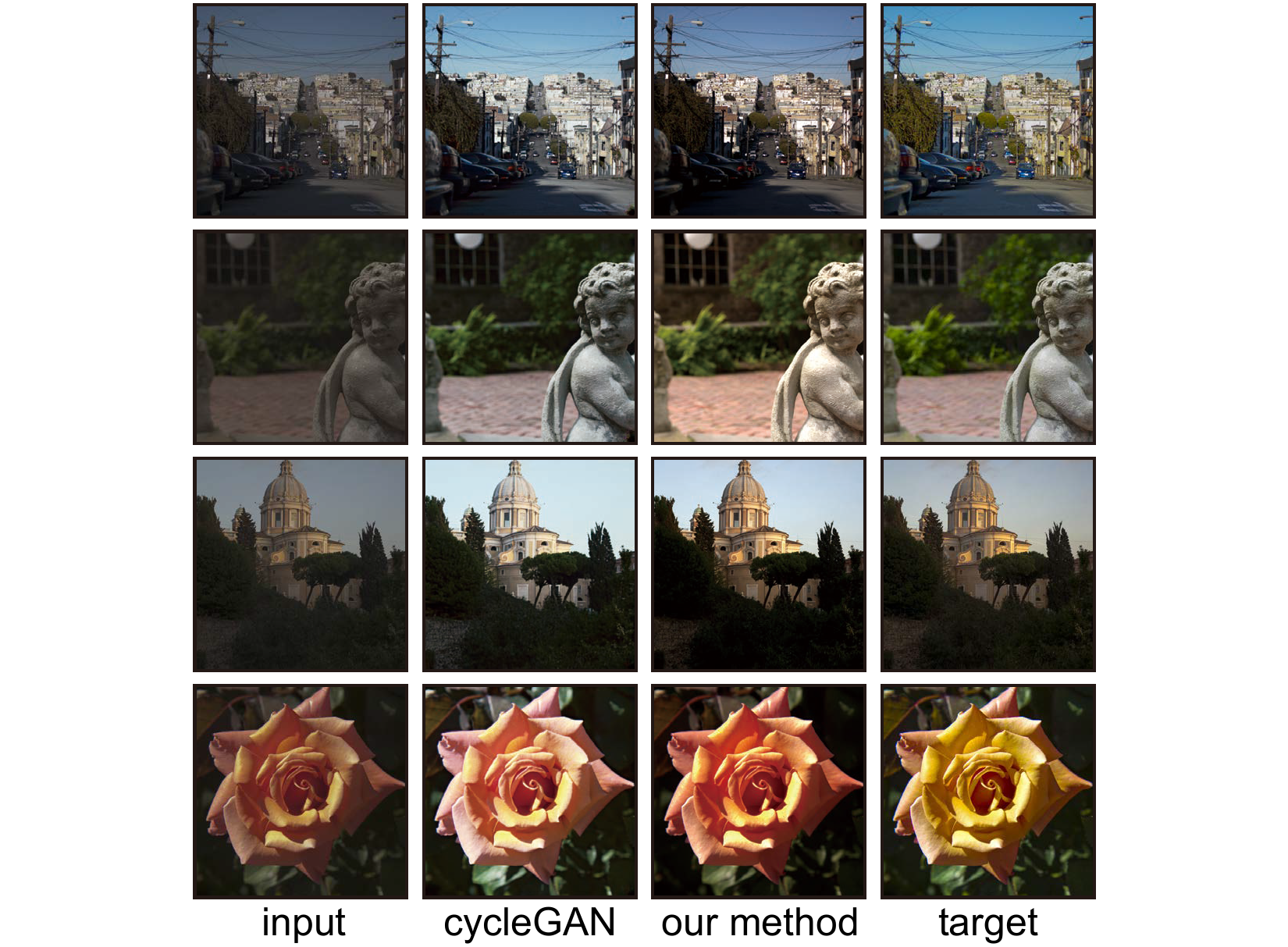}
	\caption{\textsc{Comparison with CycleGAN.} Results generated on the MIT-Adobe FiveK dataset.}
	\label{fig:compare_with_cycleGAN_fiveK}
\end{figure}

\section{CONCLUSIONS}\vspace{-0.5\baselineskip}

We have introduced a color space adaptation method for bridging the gap between synthetic and real cirrus cloud images. The adaptation method is a two-stage learning approach. A sequence of label-preserving operations is adopted in both two stages to make variants in the color space. We demonstrate that training on the adapted synthetic cirrus cloud images has significantly improved the semantic segmentation performance compared with training on the original ones.

Our method is easy to implement and deploy. It can play a primary preprocessing role in the adaptation toolset and be complementary to those convolution-based approaches. The framework is extension-friendly: other advanced color adjustment operations can be readily introduced, {\it e.g.}, the spatially variant adjustment based on curves. These operations are human-understandable and easy to control, which could improve the variability of the adversarial data for training and extend the adaptation scope of the generator. This enhanced adaptation ability is essential to adapt a more complex dataset like Virtual KITTI \cite{gaidon2016virtualworlds}. \revised{With targeted designs of label-preserving operations, our method of color space domain adaptation owns a robust performance on different tasks. Moreover, the performance improves along with the rise of the details produced by photo-realistic rendering. It is promising to apply our method to general synthetic datasets in color space and acquire proper performance improvement with the development of photo-realistic rendering technologies. }


\section*{Compliance with Ethical Standards}\vspace{-0.5\baselineskip}
Funding: This study was funded by National Natural Science Foundation of China (grant number 61772024, 61732016).

Conflict of Interest:  The authors declare that they have no conflict of interest.



%
%

\bibliographystyle{spmpsci_unsort}      
\bibliography{ref}   

\noindent \textbf{Qing Lyu}\
received the bachelor's degree in Digital Media Technology from Zhejiang University in 2016. Currently, she is working toward the PhD degree at the State Key Lab of CAD\&CG, Zhejiang University. Her research interests include visual computing and computer graphics.\par
\vspace{0.5\baselineskip}

\noindent \textbf{Minghao Chen}\
received the B.S. degree from the Zhejiang University, Hangzhou, Zhejiang, China, in 2018, where he is currently pursuing the PhD degree. His main research interests include computer vision and domain adaptation.\par
\vspace{0.5\baselineskip}

\noindent \textbf{Xiang Chen}\
is an Associate Professor in the State Key Lab of CAD\&CG, Zhejiang University. He received his Ph.D. in Computer Science from Zhejiang University in 2012. His current research interests mainly include fabrication-aware design, physics-based simulation, image analysis, shape modeling/retrieval and computer-aided design.\par

\begin{appendix}

\section*{APPENDIX}

\section{Color Adjustment Operations}\label{sec:appendix}\vspace{-0.5\baselineskip}

\paragraph{\textbf{Brightness}}
The brightness adjustment operation is defined as
\begin{equation}\label{eqn:opb}
\mathsf{op}_b(x; \alpha_b)=
\begin{cases}
  x\cdot(1-\alpha_b)+\alpha_b, \quad&\text{if}\ \alpha_b>=0\\
  x+x\cdot\alpha_b, \quad&\text{otherwise}
\end{cases}
\end{equation}
where $x$ is the input image, and $\alpha_b$ is a scalar parameter that controls the extent of the adjustment. We clip $\alpha_b$ into the range $[-1,1]$.

\vspace{-0.5\baselineskip}

\paragraph{\textbf{Saturation}}
The saturation adjustment operation is defined as
\begin{equation}
\mathsf{op}_s(x; \alpha_s)=
\begin{cases}
  x+(x-\mathsf{L}(x))\cdot\mathsf{s}(x,\alpha_s), &\text{if}\ s>0\\
  \mathsf{L}(x)+(x-\mathsf{L}(x))\cdot(1+\mathsf{s}(x,\alpha_s)), &\text{otherwise}
\end{cases}
\end{equation}
where $x$ is the input image, and $\alpha_s$ is a scalar parameter that controls the extent of the adjustment. We clip $\alpha_s$ into the range $[-1,1]$.

$\mathsf{L}(x)$ is the per-pixel average of the three channels $\frac{1}{2}\cdot[\mathsf{rgb\_max}(x) + \mathsf{rgb\_min}(x)]$, and $\mathsf{s}(x,\alpha_s)$ is defined as
\begin{equation*}
\mathsf{s}(x; \alpha_s) =
\begin{cases}
  1/\mathsf{S}(x)-1, &\text{if}\ \alpha_s+\mathsf{S}(x)>=1\\
  1/(1-\alpha_s)-1, &\text{otherwise}
\end{cases}
\end{equation*}

$\mathsf{S}(x)$ is defined as a per-pixel ratio
\begin{equation*}
\mathsf{S}(x) =
\begin{cases}
  \mathsf{delta}(x)/(2\cdot\mathsf{L}(x)), &\text{if}\ \mathsf{L}<0.5\\
  \mathsf{delta}(x)/(2-2\cdot\mathsf{L}(x)), &\text{otherwise}
\end{cases}
\end{equation*}
where $\mathsf{delta}(x) = \mathsf{rgb\_max}(x) - \mathsf{rgb\_min}(x)$.

\vspace{-0.5\baselineskip}

\paragraph{\textbf{Contrast}}
The contrast adjustment operation is defined as
\begin{equation}
\mathsf{op}_c(x; \alpha_c)=
\begin{cases}
  \bar{x}+(x-\bar{x})/(1-\alpha_c), &\text{if}\ \alpha_c>=0\\
  \bar{x}+(x-\bar{x})\cdot(1+\alpha_c), &\text{otherwise}
\end{cases}
\end{equation}
where $x$ is the input image, $\bar{x}$ is the average of all pixel values of $x$, and $\alpha_c$ is a scalar parameter that controls the extent of the adjustment. We clip $\alpha_c$ into the range $[-1,1]$.

\begin{figure*}[!t]
	\centering
	\includegraphics[width=\linewidth]{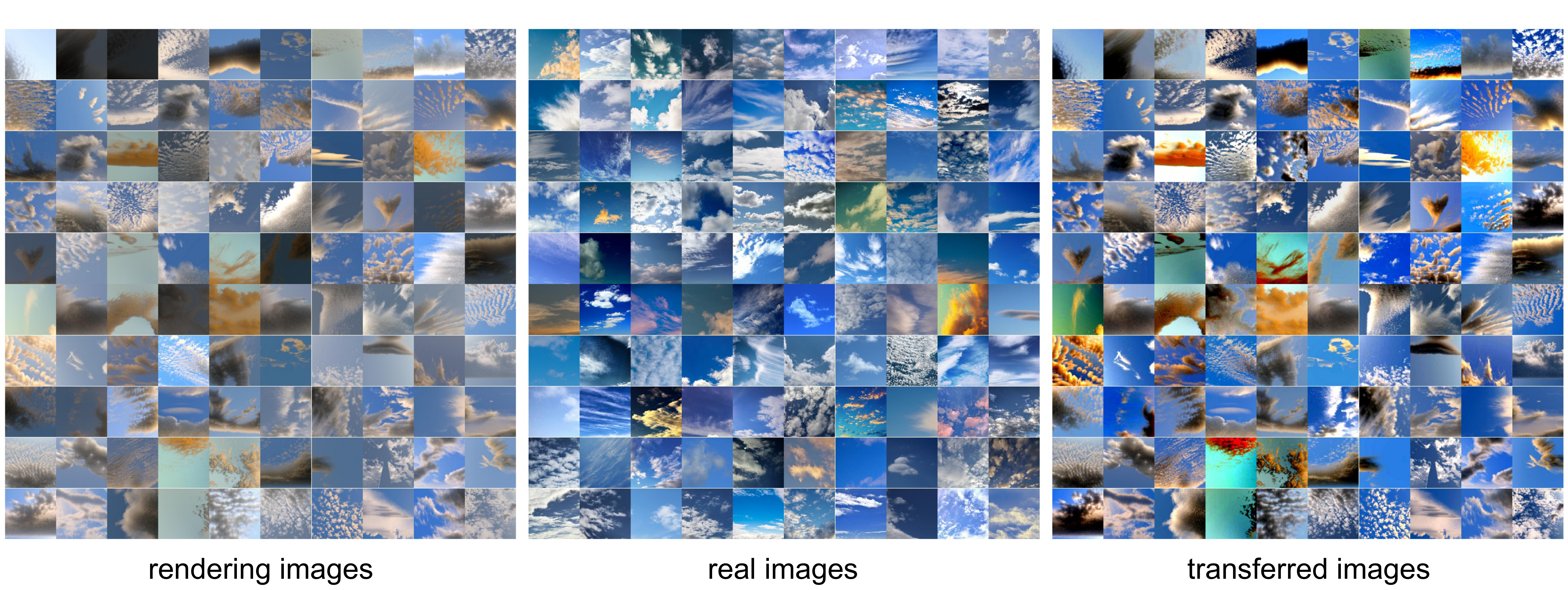}
    \caption{\textsc{More color adaptation results.}}
    \label{fig:diversity}
\end{figure*}

\vspace{-1\baselineskip}

\section{Handcrafted Approach for Image Augmentation}\label{sec:handcrafted approach}\vspace{-0.5\baselineskip}
We execute an adaptation approach using handcrafted features in the experiment. First, we transfer the images to HSV space to extract features of saturation and brightness. Then, we fit a Gaussian distribution model to the feature points of the real images. Next, for each synthetic image, we shift its features towards a target point sampled from the Gaussian distribution model. Finally, we reconstruct augmented images from the shifted features. Compared with the handcrafted feature, our generator learns more powerful features in higher dimensions and leverages that to decide the best way to shift each synthetic image.

\vspace{-1\baselineskip}

\section{Details of Style Transfer and Image Post-Processing}\label{sec:appendix_dataset}\vspace{-0.5\baselineskip}

\begin{figure}[!b]
	\centering
	\includegraphics[width=\linewidth]{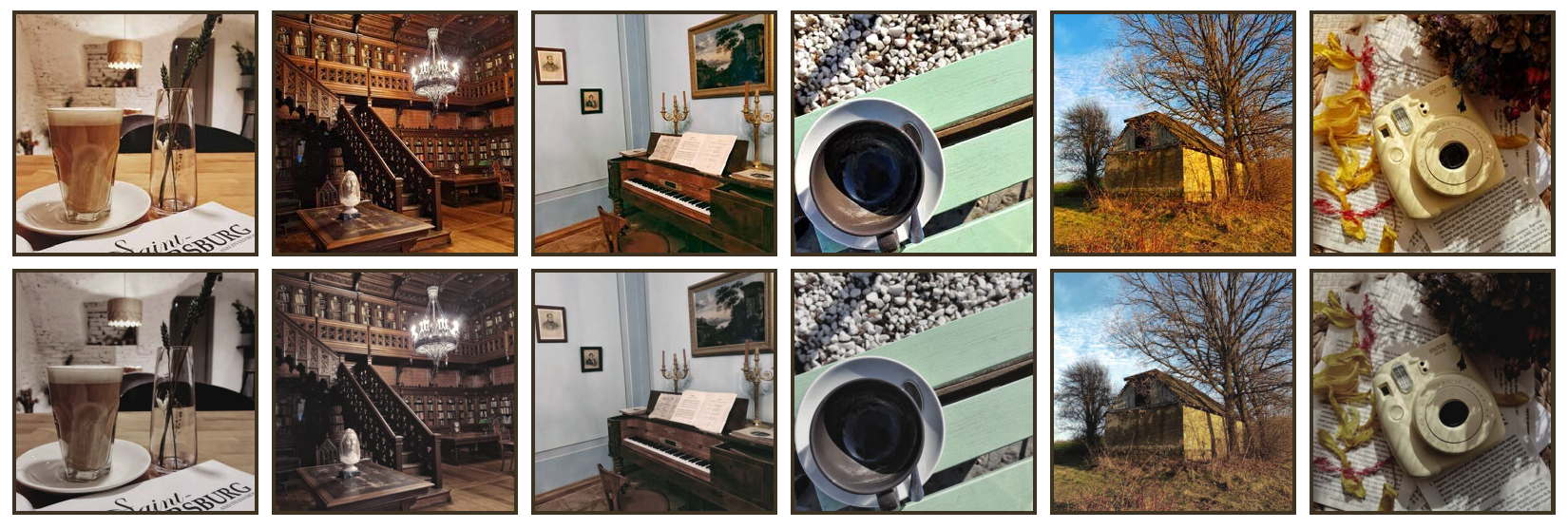}
	\caption{\textsc{Adaptation on Pexels.} Our method adapts the photos of artist A (top row) to the style of artist B (bottom row).}
	\label{fig:more_results_pexel}
\end{figure}

\begin{figure}[!b]
	\centering
	\includegraphics[width=\linewidth]{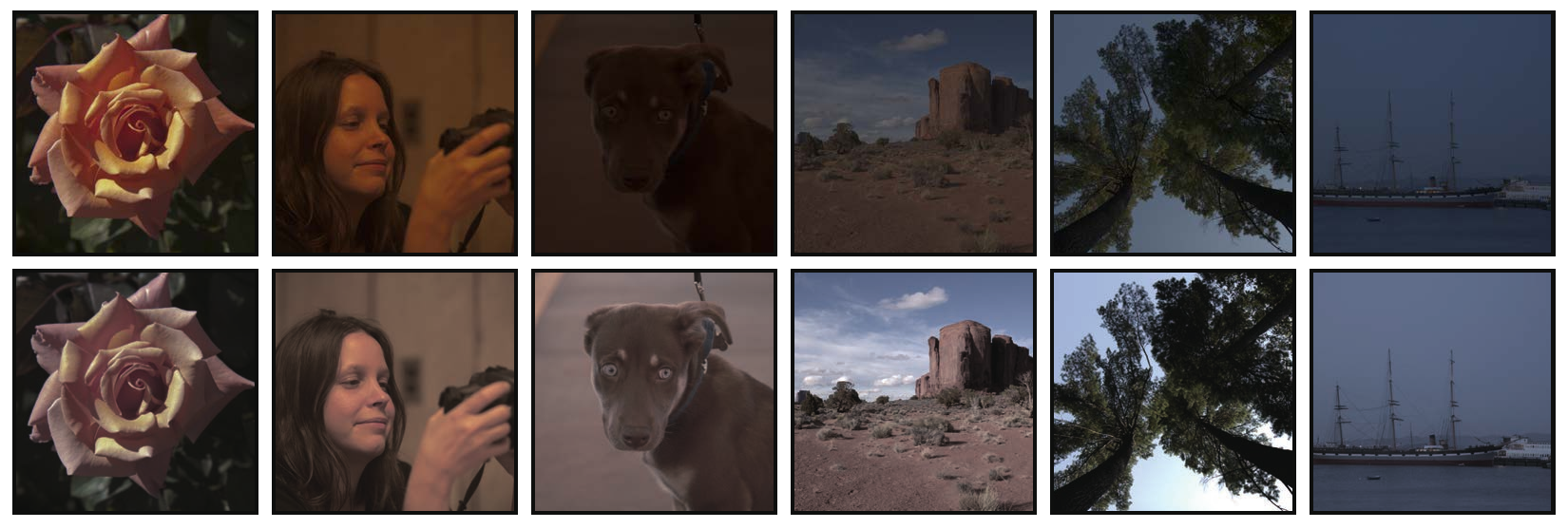}
	\caption{\textsc{Cross-dataset generalization.} The model trained on the Pexels dataset is applied to raw images of the MIT-Adobe FiveK Dataset (top row) to obtain an artist style (bottom row).}
	\label{fig:apply_on_other_dataset}
\end{figure}

All the training images are zero-centered and rescaled to $[-1, 1]$. We set the batch size to 8. We adopt the Adam optimizer with $lr=2\mathrm{e}{-4}$ and $\beta_1=0.5$, and train both the discriminator and the generator for 100 epochs. We show more color adaptation results on the Pexels dataset in Fig.~\ref{fig:more_results_pexel}. We also apply the model trained on the Pexels dataset to images in the MIT-FiveK dataset to show the ability of cross-dataset generalization (Fig.~\ref{fig:apply_on_other_dataset}). While similar effects can be produced by \cite{hu2018exposure}, our method does not require reinforcement learning.

\begin{figure*}[!t]
	\centering
	\includegraphics[width=0.7\linewidth]{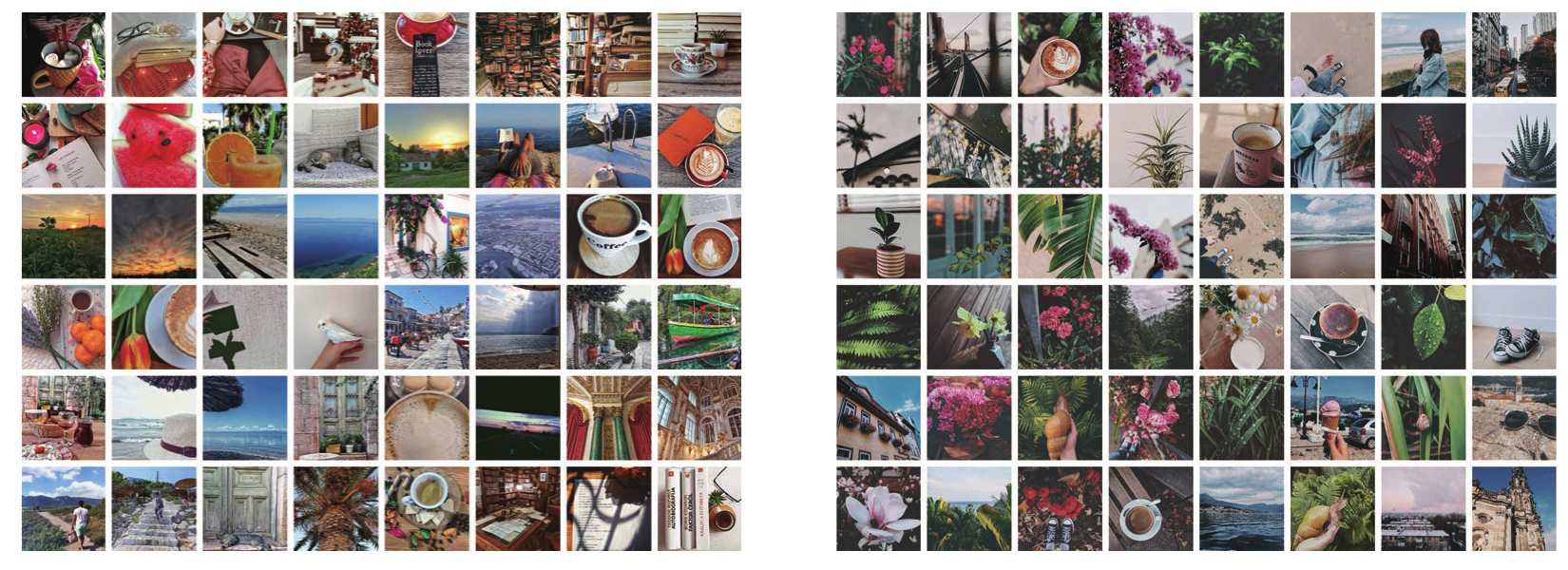}
	\caption{\textsc{Styles overview.} Retouched images of artist A (left) and artist B (right) from \emph{pexels.com}.}
	\label{fig:dataset}
\end{figure*}

\end{appendix}

\end{document}